%% file: ark2012-template.tex
\documentclass{svmult}

\usepackage{graphics} 
\usepackage{epsfig} 
\usepackage{psfrag}
\usepackage{subfigure}
\usepackage{helvet}
\usepackage{courier}
\usepackage{amsmath,amsfonts}
\usepackage{graphicx}
\usepackage[bottom]{footmisc}
\usepackage{tabularx}
\usepackage{amsmath,amssymb}
\usepackage[dvips]{floatflt}

\newcommand{\MVC}{\operatorname{MVC}}

\newcommand{\cem}{\operatorname{cem}}
\newcommand{\shoulder}{\operatorname{shoulder}}
\newcommand{\elbow}{\operatorname{elbow}}

\newcommand{\joint}{\operatorname{joint}}
\newcommand{\equals}{\stackrel{\mathrm{def}}{=}}

\begin{document}
\title*{Human Muscle Fatigue Model in Dynamic Motions}
\author{Ruina Ma \and Damien Chablat \and Fouad Bennis \and Liang Ma}
\institute{%
  Ruina Ma, Damien Chablat, Fouad Bennis, \at
  Institut de Recherche en Communications et Cybern\'etique de Nantes, UMR CNRS 6597 
  \email{\{Ruina.Ma, Damien.Chablat, Fouad.Bennis\}@irccyn.ec-nantes.fr}
  \\
  Liang MA, \at Department of Industrial Engineering, Tsinghua University. \goodbreak          
  \email{liangma@tsinghua.edu.cn}}

\maketitle
\vspace{-10mm}
\abstract{Human muscle fatigue is considered to be one of the main reasons for Musculoskeletal Disorder (MSD). Recent models have been introduced to define muscle fatigue for static postures. However, the main drawbacks of these models are that the dynamic effect of the human and the external load are not taken into account. In this paper, each human joint is assumed to be controlled by two muscle groups to generate motions such as push/pull. The joint torques are computed using Lagrange's formulation to evaluate the dynamic factors of the muscle fatigue model. An experiment is defined to validate this assumption and the result for one person confirms its feasibility. The evaluation of this model can predict the fatigue and MSD risk in industry production quickly.
}
\vspace{-5mm}
\input{Introduction}
\vspace{-5mm}
\input{FatigueModelExtend}
\vspace{-5mm}
\input{Experiment}
\vspace{-5mm}
\input{Case_Study_K_evaluation}
\vspace{-5mm}
\input{Discussion}
\vspace{-5mm}
 \bibliographystyle{spmpsci}
 \bibliography{Bibilography}
\end{document}

%% file: Introduction.tex
\section{Introduction}

Muscle fatigue is defined as ``any reduction in the ability to exert force in response to voluntary effort''~\cite{Chaffin1999} and is one of the main reasons leading to MSD~\cite{Kumar2008}. From Hill's muscle model~\cite{Hill1983} to today's muscle fatigue models, this topic has been researched from different scientific field with special point of views. In general, mainly two approaches have been adopted to evaluate muscle fatigue~\cite{Westgaard2000}, either in theoretical methods or in empirical methods. In~\cite{Wexler1997}, Wexleret et al. proposed a new muscle fatigue model based on Ca\textsuperscript{2+} cross-bridge mechanism and verified the model with simulation experiments. Although this model can be used to predict the muscle force fatigue under different simulation frequencies, the large number of variables make it difficult to use compare with other models. In~\cite{Liu2002}, Liu et al. proposed a fatigue and recovery models based on motor units pattern. They demonstrated the relationship among muscle activation, fatigue and recovery. This model is available under maximum voluntary contraction situation; this condition is rare in the manual working situation. Another muscle fatigue model was developed by Giat~\cite{Giat1993} based on force-pH relationship. This fatigue model was obtained by curve fitting of the pH level with time in the course of stimulation and recovery, but it cannot used in evaluating the muscle fatigue in the whole working process. In~\cite{Ma2009Thesis}, Ma et al. proposed a muscle fatigue model from the macroscopic point of view. External physical factors and personal factors were taken into consideration to construct the model. This model can predict the muscle fatigue trend in static working posture ($\theta_{\elbow} = 90^\circ$, $\theta_{\shoulder} = 30^\circ$), but in dynamic working situation this model was limited. 

The purpose of this work is to extend muscle fatigue model to dynamic working situations. The difference force generation between static working posture and dynamic working motions is depended on the activation of different muscle types. There are three types of fibers of muscle: slow-twitch fibers, fast-twitch A fibers and fast-twitch B fibers~\cite{Karp2001}. In every postures and motions all of the three muscle fibers are used, but the percent of every fibers in static and dynamic situation is different. In a static working posture fast-twitch fibers is mostly used and this type of muscle fibers have a low resistance to fatigue. In a low speed dynamic working motions slow-twitch fibers are mainly used and this type of muscle fibers have a high resistance to fatigue. Meanwhile, the blood circulation during dynamic motions is better than in a static working posture. For these reasons, the behavior of the muscle and its fatigue rate are different in the two types of situations. In this paper, a muscle fatigue model in dynamic situation is proposed. A new approach to identify the fatigue rate parameter $k$ is used. An experimental setup is defined to validate this assumption.
 
Firstly, some assumptions are given and a new dynamic muscle fatigue model is proposed. Secondly, an experiment is designed to verify this model. Thirdly, a case-study for one person is illustrated and the fatigue parameter $k$ is evaluated. Finally, some perspectives are presented.

%% file: FatigueModelExtend.tex
\section{Proposal of a new muscle fatigue model}

\noindent \underline {Dynamic muscle fatigue model:} Muscles in the human body have one most important function, such as force generating devices. They can work only in a single direction. Hence, for each single joint, at least two groups of muscles (agonistic muscle and antagonistic muscle) are necessary to control the motion. The co-contraction of the two groups of muscles provide stability to joint and balance to the posture. From the articulation point of view, it is assumed that a joint is controlled by two groups of muscles (one for flexion and one for extension). These muscle groups create a torque on the joint. This torque drives the human movement and whether it is positive or negative depends on the angle and direction of joint rotation. Based on the previous model of Ma et al., we propose that:
\begin{enumerate}
	\item the fatigue of muscle is proportional to the joint torque, i.e. in the same period of time, the larger the torque of joint exerted, the more fatigue people feel;
	\item the fatigue of muscle is inversely proportional to the muscle torque capacity i.e. the smaller the capacity is, the quicker the muscle becomes tired.
\end{enumerate}
This can be mathematically described by the following equation.
\begin{equation}
	\frac{d \Gamma_{\cem}(t)}{dt} = -k\ \cdot\ \frac{\Gamma_{\cem}(t)}{\Gamma_{\MVC}} \cdot \Gamma_{\joint}(t)
	\label{eq:fatigue4}
\end{equation}
where the set of parameters are listed in Table~\ref{tab:DynamicFatigueParas}.
\vspace{-5mm}
\begin{table}[h!]
\centering
\begin{tabularx}{0.95\linewidth}{|l|l|X|}
	\hline
	Item & Unit & Description\\
	\hline
	$\Gamma_{\MVC}$ 		& N.m &	Maximum voluntary contraction of joint torque, i.e. $\Gamma_{\max}$\\
	\hline
	$\Gamma_{\cem}(t)$ 	& N.m & Current exertable maximum joint torque\\
	\hline
	$\Gamma_{\joint}$	  & N.m & Joint torque, i.e. the torque which the joint needs to generate\\
	\hline
	$k$				       		& $\min^{-1}$ & Fatigue rate\\
		\hline
\end{tabularx}
\caption{\label{tab:DynamicFatigueParas}Parameters in dynamic fatigue model}
\end{table}
\vspace{-5mm}

\noindent If we assume that $\Gamma_{\cem}(0)=\Gamma_{\MVC}$ and $k$ is a constant, the integration result of the previous equation is given by
\begin{equation}
	\Gamma_{\cem}(t) = \Gamma_{\MVC}\ \cdot\ e^{ - \frac{k}{\Gamma_{\MVC}} \int_{0}^{t} \Gamma_{\joint}(u) du}
	\label{eq:fatigue5}
\end{equation}
The value of $\Gamma_{\MVC}$ is a fixed value determined by individual person. In the first approximation, we assume that $\Gamma_{\MVC}$ is a constant of a joint torque during a limited period of time. According to robotic dynamic model~\cite{Khalil2010}, $\Gamma_{\joint}(u)$ can be modeled by a variable depending on the angle, the velocity, the acceleration and the internal/external load.
\begin{equation}
	\Gamma_{\joint}(u) \equals \Gamma (u, \theta, \dot{\theta}, \ddot{\theta})
\end{equation}
This way, Equation~\eqref{eq:fatigue5} can be further simplified in the form.
\begin{equation}
	\Gamma_{\cem}(t) = \Gamma_{\MVC}\ \cdot\ e^{ -\frac{k}{\Gamma_{\MVC}} \int_{0}^{t}\Gamma (u, \theta, \dot{\theta}, \ddot{\theta}) du} 
	\label{eq:fatigue6}
\end{equation}
Equation~\eqref{eq:fatigue6} defines our new dynamic muscle fatigue model. The model takes consideration of the motion by the variations of the torque $\Gamma_{\joint}$ from joint level. This torque which is computed using robotic method is integrated to obtain the current exertable maximum joint torque. At first stage, we do not take into account the muscle co-contraction factor~\cite{AitHaddou2000}. This work enlarges the muscle fatigue model usefulness range.

The new dynamic fatigue model is in joint level. As mentioned in the assumption, the motion of joint is driven by a pair of muscles. Obviously, this model can be easily applicated in muscle level. For one cycle, the $\Gamma_{\joint}$ is negative or positive related with elbow rotation range. If $\Gamma_{\joint}$ is positive, we suppose it is the effect of agonistic muscle. Inversely, if $\Gamma_{\joint}$ is negative, we suppose it is the effect of antagonistic muscle. Based on this consideration, there will be two fatigue rate parameters $k$ ($k_{\textrm{agonist}}$ and $k_{\textrm{antagonist}}$) for a dynamic operation in the muscle level. Our new dynamic fatigue model can applicate fatigue evaluation in muscle level.

%% file: Experiment.tex
\section{\label{sec:experimentation}Experiment design for validation}

\begin{floatingfigure}[r]{45mm}
	\centering
	{\includegraphics[width=45mm]{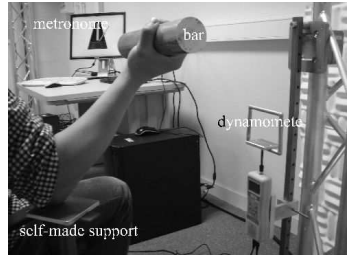}}
	\caption{\label{fig:Experiment_schema} Measurement device }
\end{floatingfigure}
The aim of the experiment design is to evaluate the muscle fatigue model. We suppose that in a push operation the agonistic muscle is mainly used whereas in a pull operation the antagonistic muscle is mainly used. Based on this assumptions, we concentrate the study on the elbow joint and use a push/pull operation to simulate dynamic motions. This evaluation consists in measuring the maximal push and pull strength after a continuous movement of the lower arm.\\

\noindent \underline {Experiment materials (Figure~\ref{fig:Experiment_schema}):}
\begin{enumerate}
	\item \emph{A dynamometer.} This device is used to measure the maximum push/pull force after lengths of time's movement of the lower arm.
	\item \emph{A bar.} This weight is grabbed by the hand of the participant and is used to simulate the weight of an operation tool in industrial environment. For our experiment the bar weight is 3 Kg.
	\item \emph{A metronome.} This tool is used to define the sample times of the motion. For our experiment the frequency is 1Hz.
	\item \emph{A self-made support.} This support is used to maintain the elbow posture during the motion and measure the torque after the operation.
\end{enumerate}

\noindent \underline {Experiment procedure:} The participant seats in a chair and puts his elbow on the support. The procedure is to repeat a rotation of the elbow joint from 0 to 75 degrees and then from 75 degrees to 0 during $t_i$ unit of time. This movement is done with the bar in hand.
The experiment procedure is as follows:
\begin{enumerate}
	\item Measure $\Gamma_{\MVC}$ before starting the operation ($\Gamma_{\MVC}(0) = \Gamma_{\cem}(0)$);
	\item Perform the dynamic operation during $t_{i}$ unit of time;
	\item Measure the remained maximum torque of elbow joint $\Gamma_{\cem}(t_i)$;
	\item Take a rest about 1-2 hours until complete recovery;
	\item Repeat steps 2, 3 and 4 for different values of $t_i$, $t_i \in \{ 0, 1, 2, 3, 4, 5 \}$ minutes.
\end{enumerate}

%% file: Case_Study_K_evaluation.tex
\section{\label{sec:case_study}Case-study of muscle fatigue for the elbow joint}

\underline{Kinematics and dynamics of the arm:}
In this section we will illustrate the dynamic calculation in our dynamic muscle fatigue model using robotic method. An example of the lower arm cyclic periodic movement during 2 seconds is demonstrated in details.\\

\begin{figure}[htb]
	\centering
	{\includegraphics[width=0.26\textwidth]{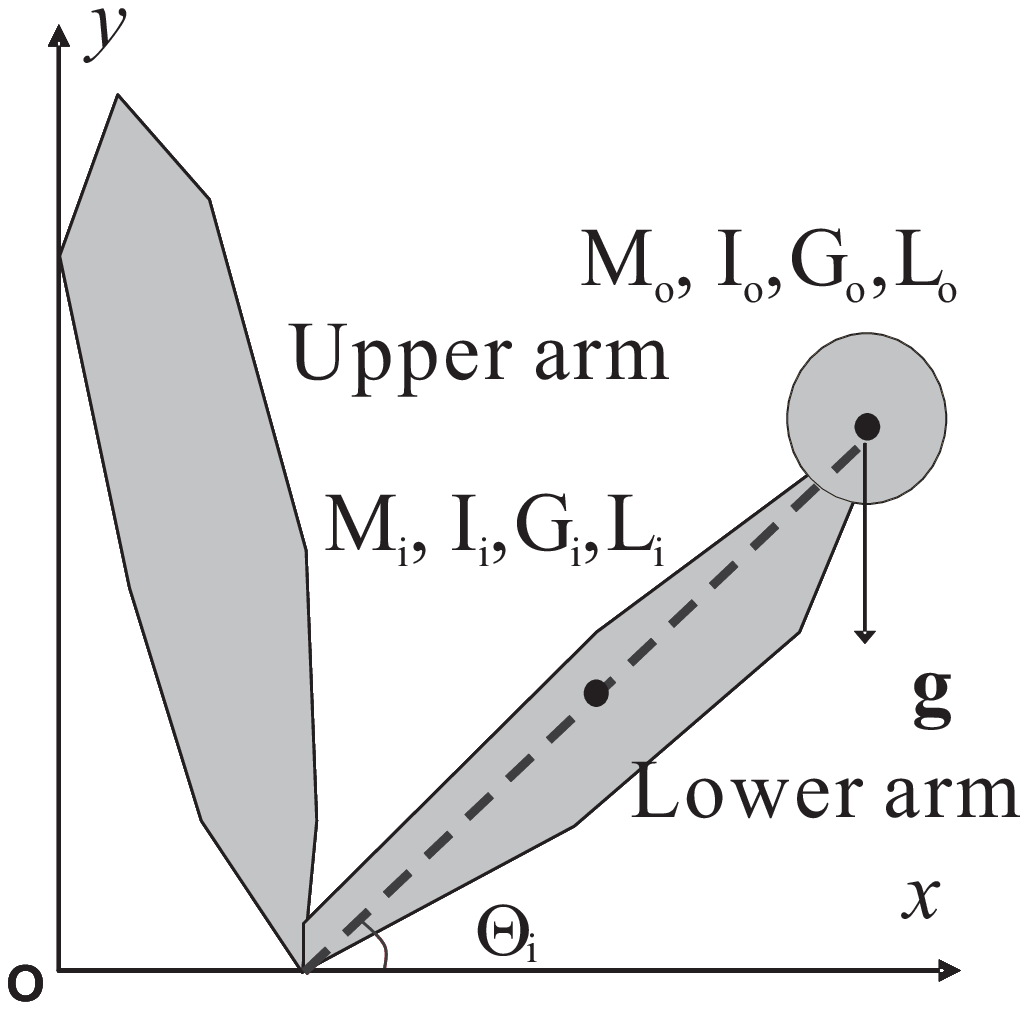}}{\small (a)}
	{\includegraphics[width=0.26\textwidth]{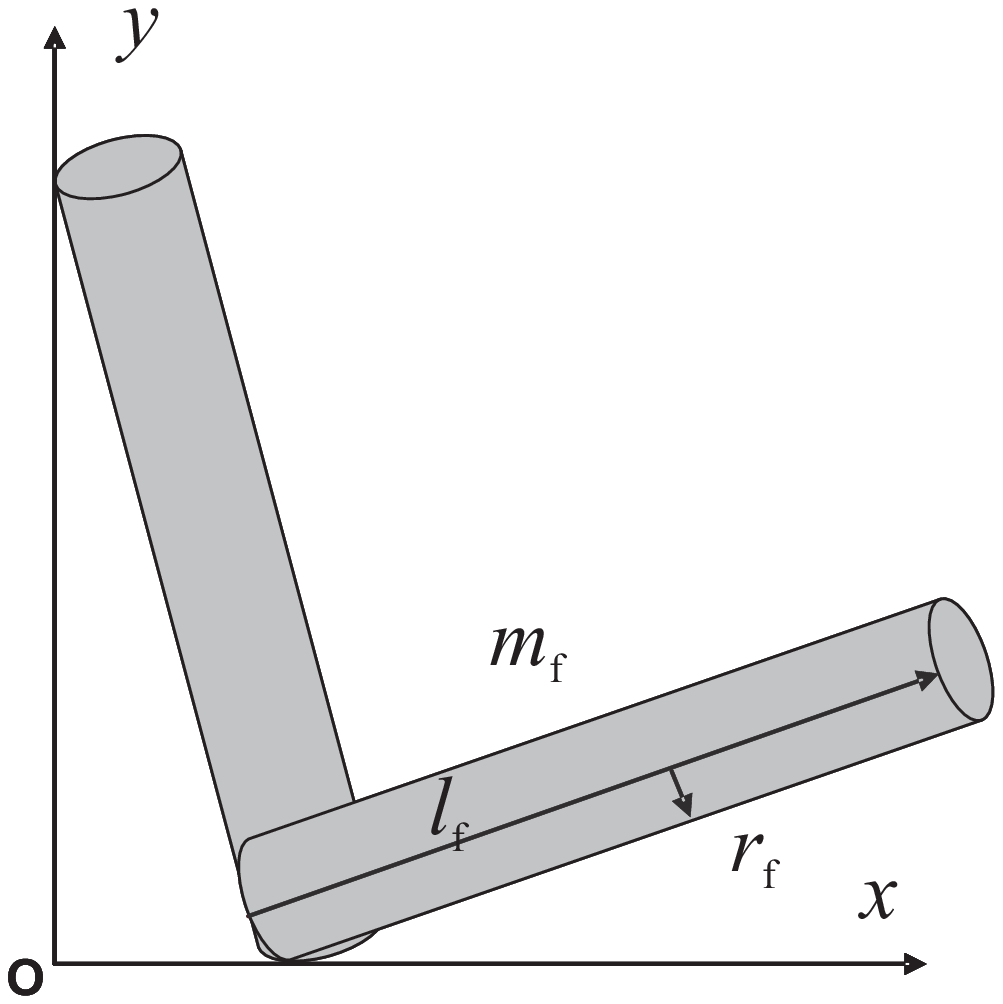}}{\small (b)}
	\caption{\label{fig:Arm_Dynamic}(a) Geometric and inertia parameters of one arm, (b) Simplified arm structure }
\end{figure}

\vspace{-5mm}
\noindent \textbf{Geometric modeling of arm: }
As presented in Figure~\ref{fig:Arm_Dynamic}, the arm model is composed of an upper arm, a lower arm and a hand. Figure~\ref{fig:Arm_Dynamic}(b) gives the simplified model of the arm used for the calculation. 
In order to determine the torque $\Gamma_{\joint}$, several parameters of the arm need to be obtained. These parameters are the length of the lower arm ($\ell_\textrm{f}$), the length of the hand ($\ell_\textrm{h}$), the radius of the lower arm ($r_\textrm{f}$) and the mass of the lower arm ($m_\textrm{f}$). If the human has a height of $H$ and a weight of $M$, according to the anthropometry database~\cite{Chaffin1999}, the related geometric human parameters are: $\ell_{\textrm{f}} = 0.146 H, \quad r_{\textrm{f}} = 0.125  \ell_{\textrm{f}}, \quad \ell_{\textrm{h}} = 0.108  H, \quad m_{\textrm{f}} = 0.023 M$.
\vspace{-5mm}

\begin{figure}[htb]
   \centering
   {\includegraphics[scale=0.24]{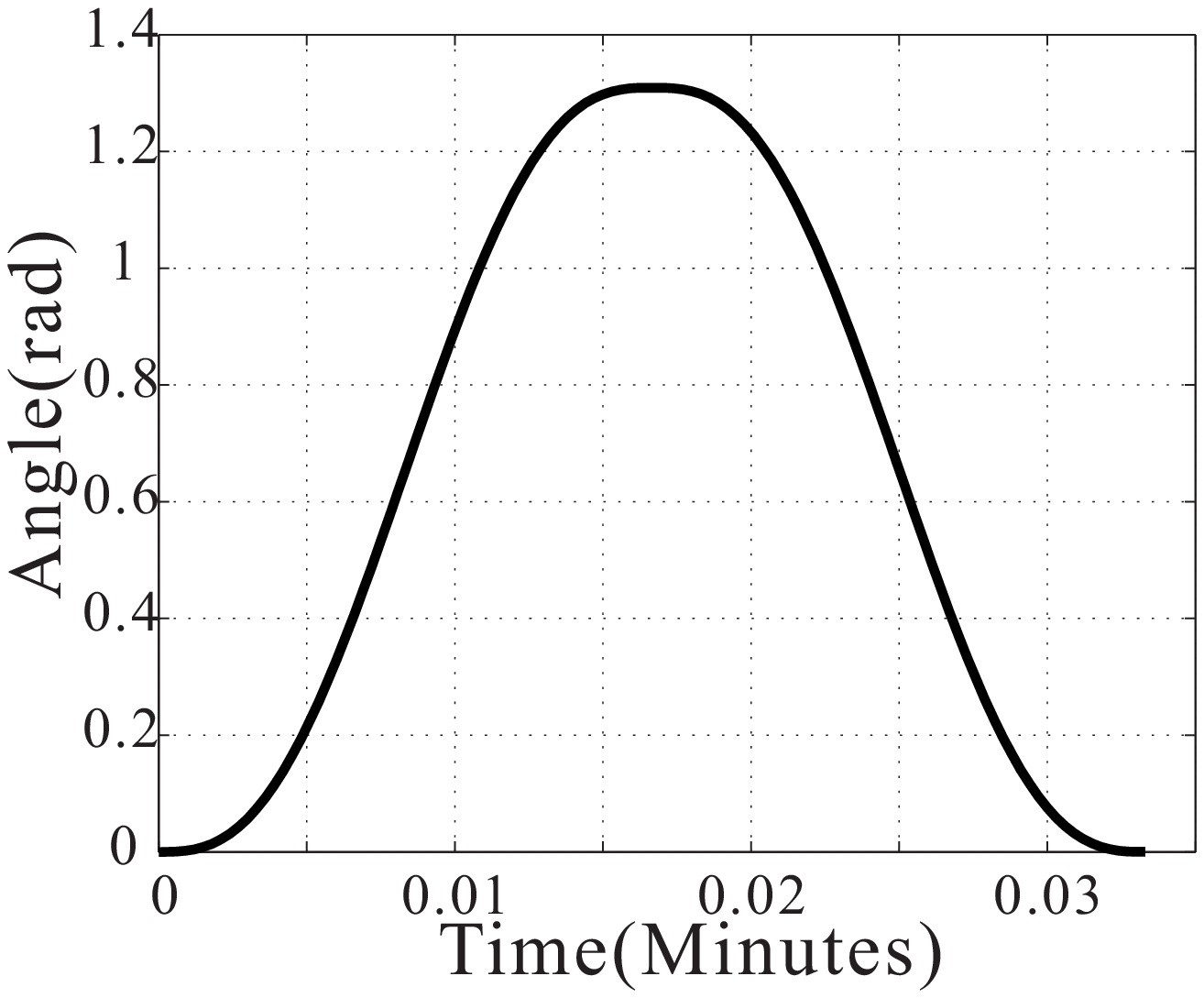}}{\small (a)}
   {\includegraphics[scale=0.24]{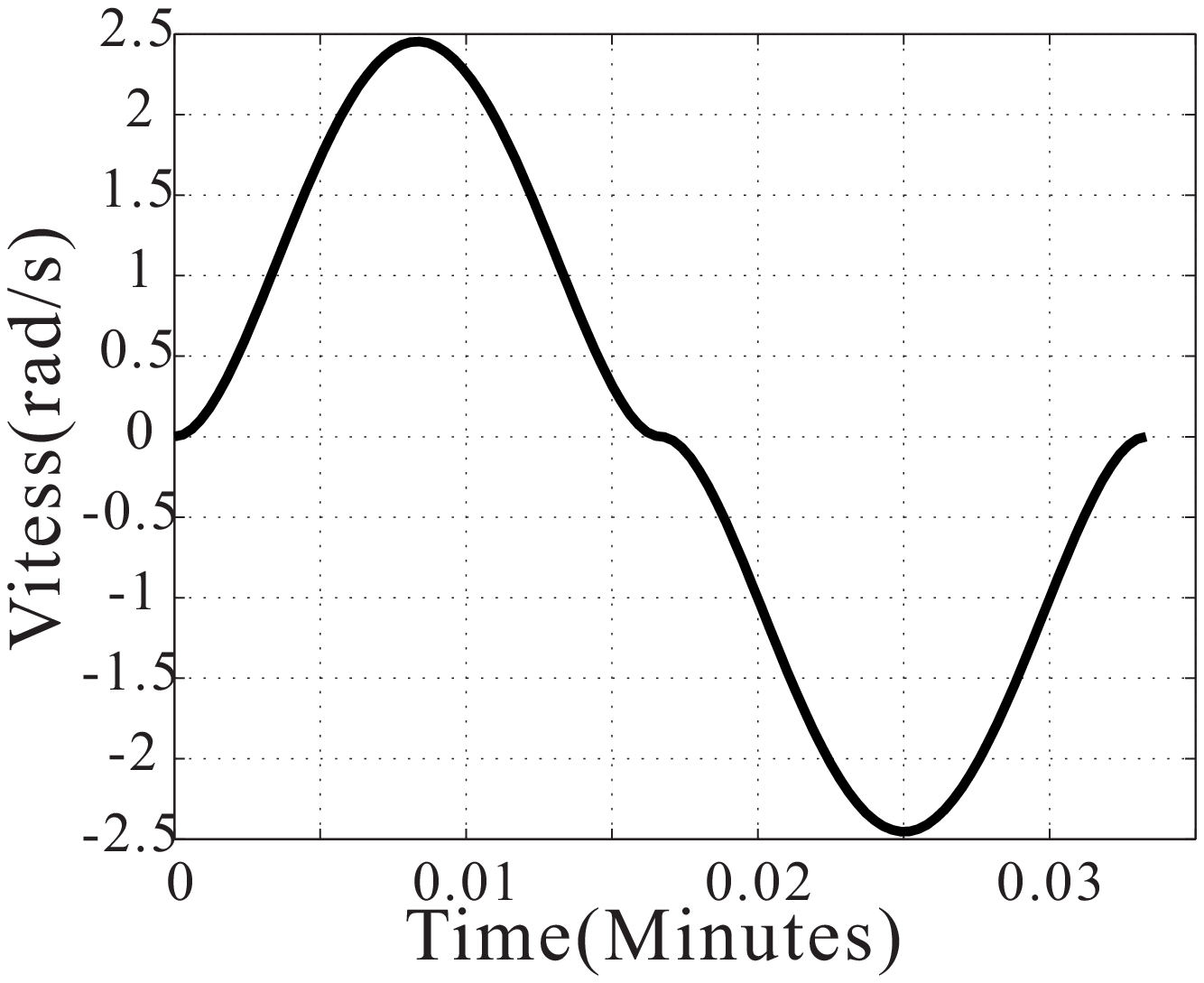}}{\small (b)}
   {\includegraphics[scale=0.24]{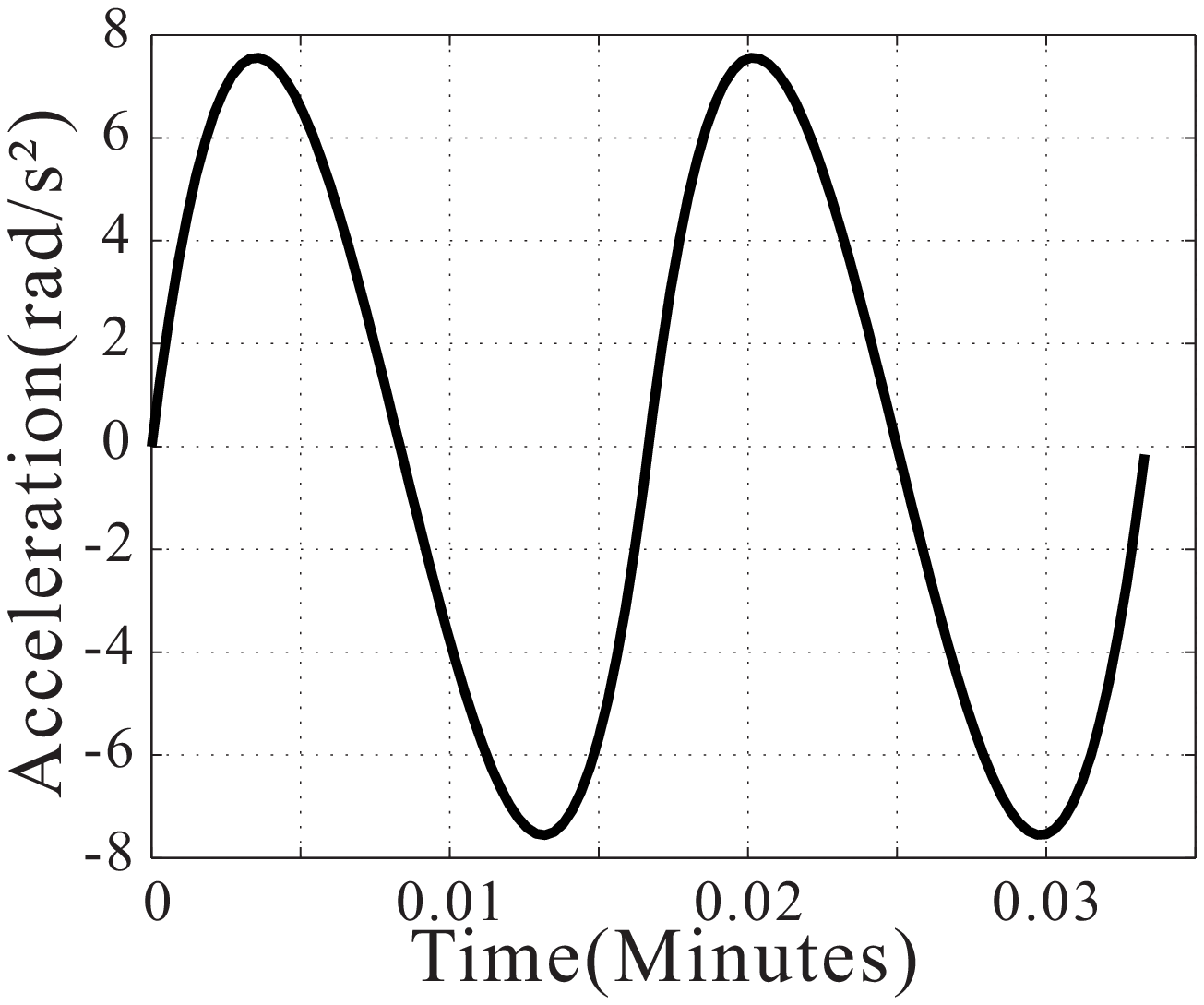}}{\small (c)}
   \caption{\label{fig:trajectory}(a) Joint angle, (b) Joint angular velocity, (c) Joint angular acceleration evolution during one cycle motion}
\end{figure}

\vspace{-5mm}
\noindent \textbf{\label{sec:trajectory_generation}Trajectory generation: }
We suppose that for one motion of lower arm up and down movement both the initial and the final velocity and acceleration are nul. We use a polynomial function to describe this movement. According to the hypothesis, the minimum degree of the polynomial satisfying the constraints is at least five and has the following form.
\begin{equation}
  P = a_0 + a_1 t + a_2 t^{2} + a_3 t^{3} + a_4 t^{4} + a_5 t^{5}
	\label{eq:polynomial}
\end{equation} 
where the coefficients $a_i$ are determined from the boundary conditions:
\begin{equation}
\begin{array}{llllll}
	\theta (0)   &= \theta^{initial}, & \dot{\theta}(0) 	&= 0, & \ddot{\theta}(0)   &= 0\\
	\theta (t_f) &= \theta^{end}, 	 	& \dot{\theta}(t_f) &= 0, & \ddot{\theta}(t_f) &= 0
\end{array} 
\label{eq:initial_condition}
\end{equation}
The trajectory between $\theta^{\textrm{initial}}$ and $\theta^{\textrm{end}}$ is determined by
\begin{equation}
	 \theta(t) = \theta^{initial} + r(t) \cdot \left( \theta^{\textrm{end}} - \theta^{\textrm{initial}} \right), \quad 0\leq t \leq t_f
	\label{eq:Joint_Position}
\end{equation}
Solving Equation~\eqref{eq:Joint_Position} with the above mentioned condition we can get the following interpolation function
\begin{equation}
	r ( t ) = 10 \left( t / t_f \right)^3 - 15 \left( t / t_f \right)^4 + 6 \left( t / t_f \right)^5
	\label{eq:Interpolation}
\end{equation}
Based on this interpolation function we can get the velocity and acceleration of every moment in the joint trajectory. The Figure~\ref{fig:trajectory} represents the evolution of $\theta$, $\dot{\theta}$ and $\ddot{\theta}$ for the considering experiment with the angle change between 0 to $5\pi/12$.\\
\vspace{-5mm}

\vspace{-5mm}
\noindent \textbf{Dynamic model and joint torque evaluation:} 
The Lagrange method is applied to compute the dynamic model~\cite{Khalil2010}.
Firstly, we calculate the joint kinetic energy and the joint potential energy
\begin{equation}
	E= E_{\textrm{joint}} + E_{\textrm{object}},  \quad U= U_{\textrm{joint}} + U_{\textrm{object}}
  \label{eq:Kinetic}
\end{equation}
Then, the joint torque is given by 
\begin{equation}
	\Gamma = \frac{d}{dt} \left( \frac{\partial L}{\partial \dot{\theta}} \right) - \left( \frac{\partial L}{\partial \theta} \right)
	\label{eq:JointTorque}
\end{equation}
where $L = E - U$.
\begin{figure}[ht!]
   \centering
   {\includegraphics[scale=0.24]{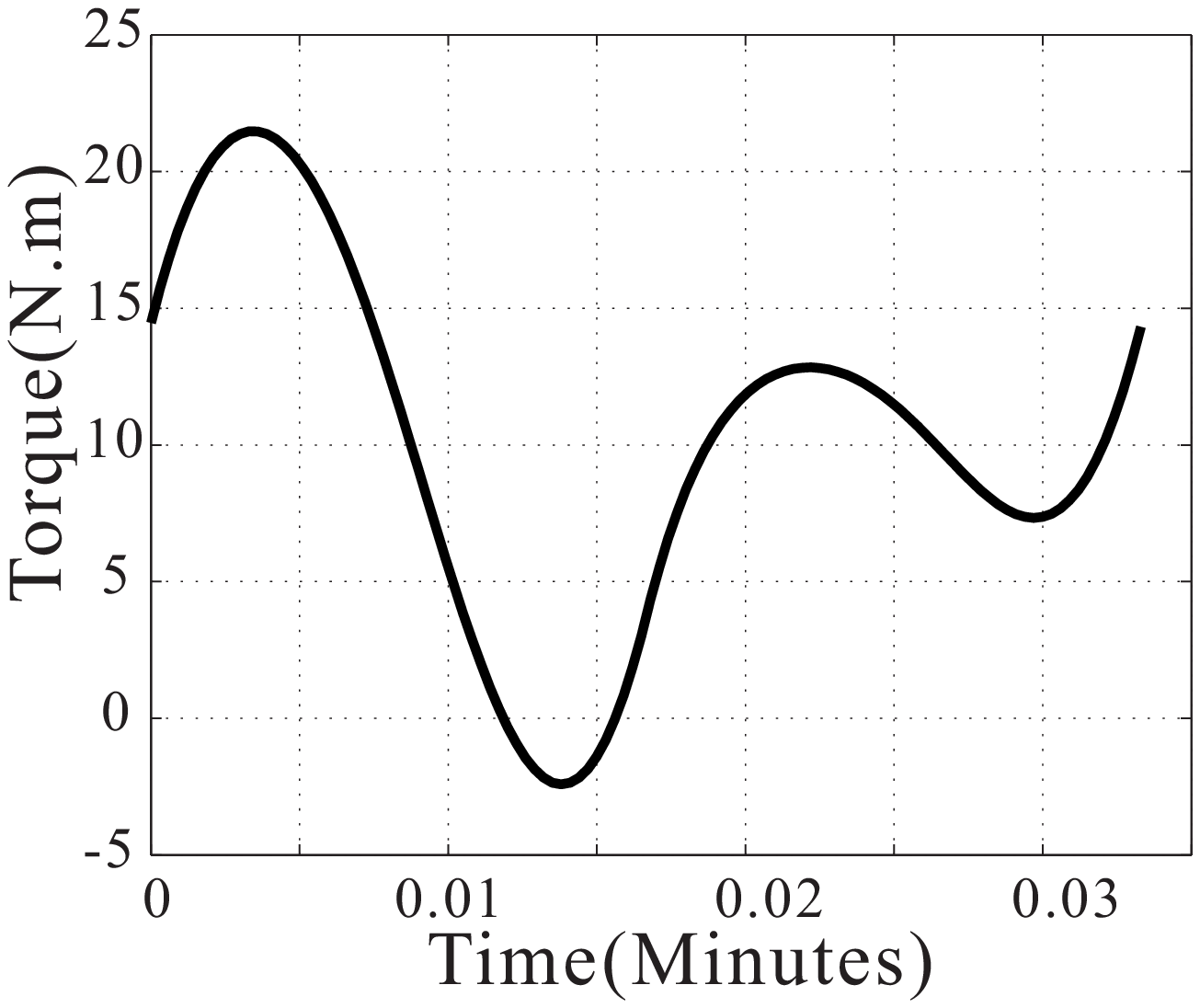}}{\small (a)}
   {\includegraphics[scale=0.24]{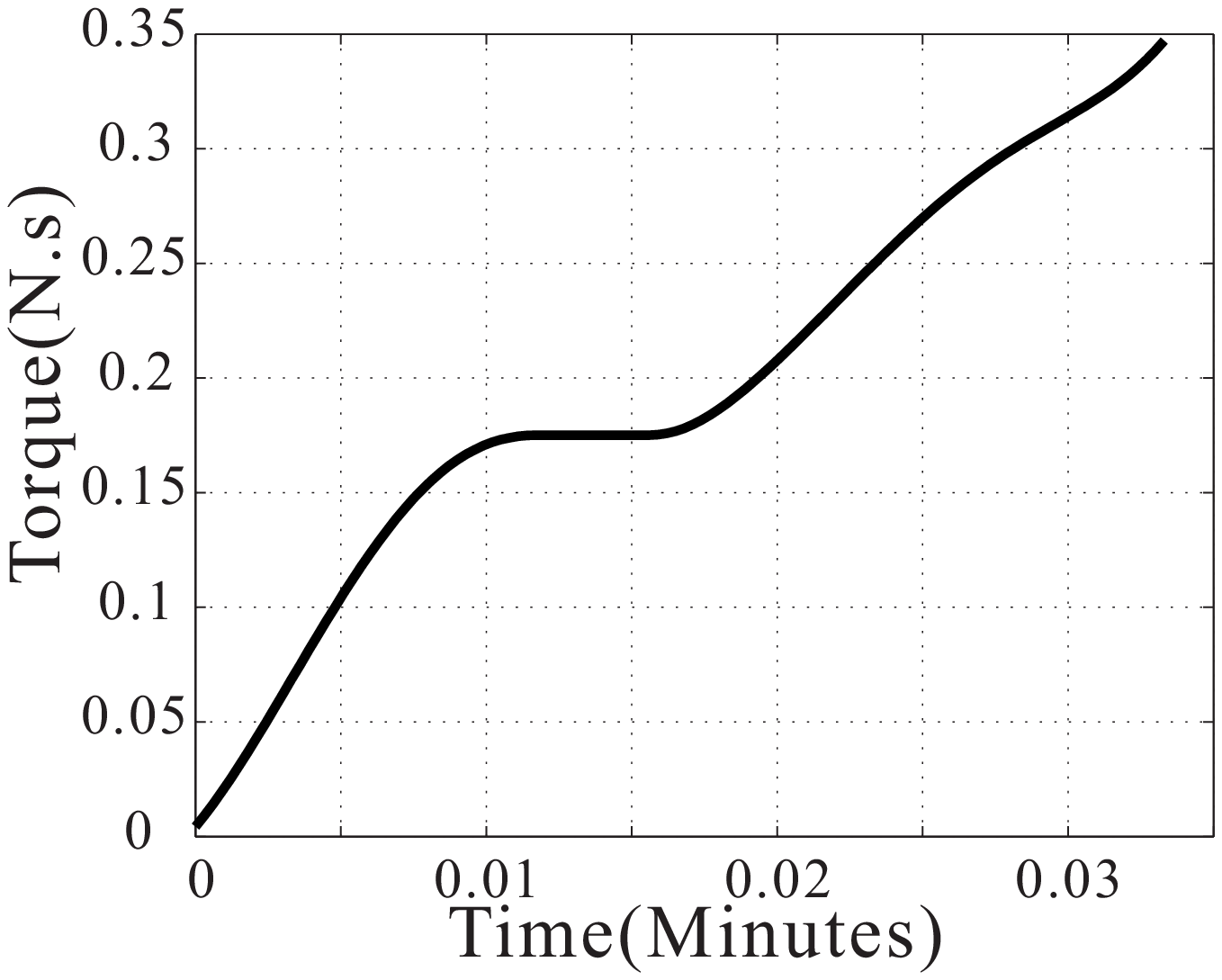}}{\small (b)}
   {\includegraphics[scale=0.24]{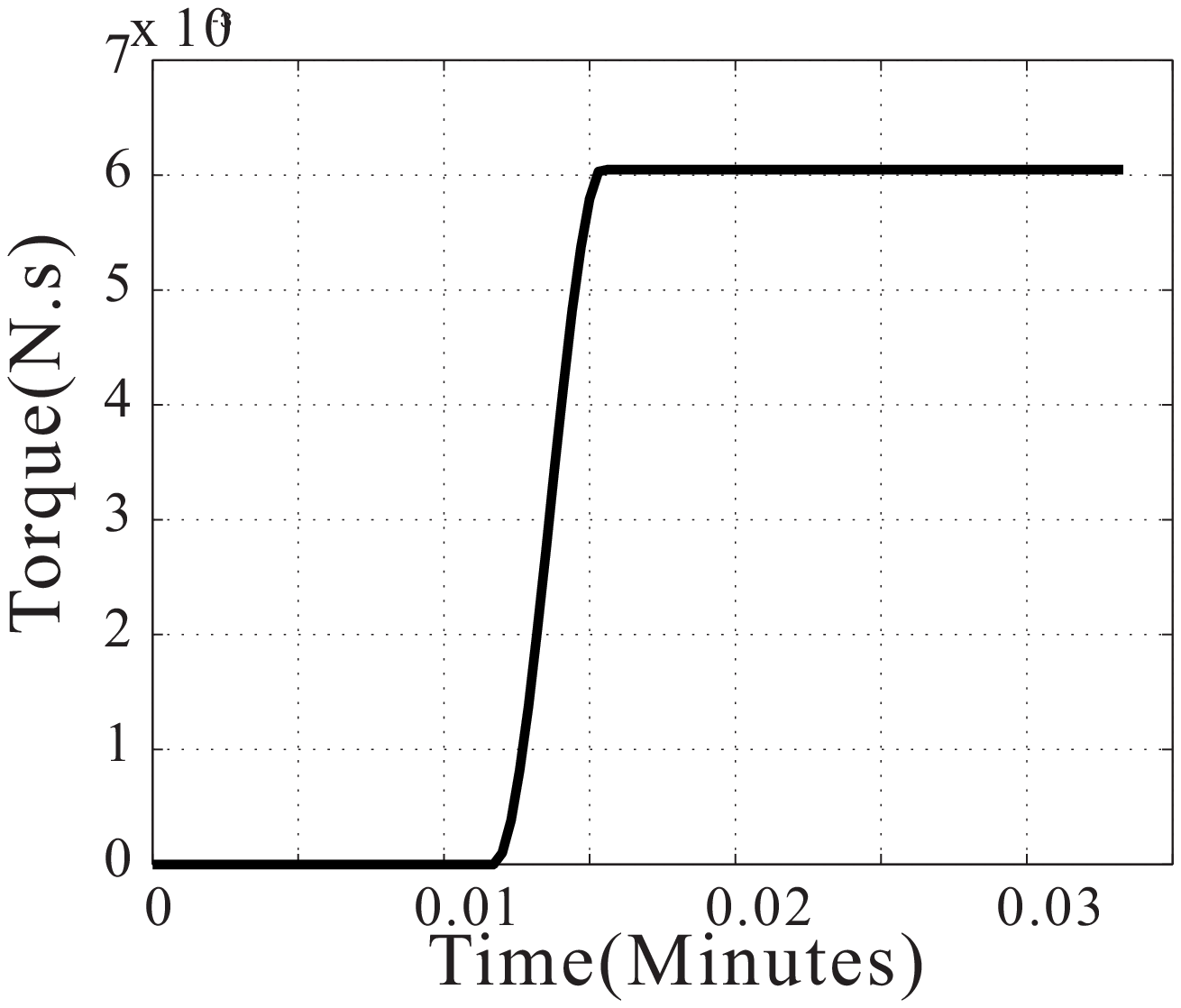}}{\small (c)}
   \caption{\label{fig:Momentum}(a) Elbow joint torque, (b) Momentum of agonistic muscle, (c) Momentum of antagonistic muscle}
\end{figure}
\vspace{-5mm}
In our dynamic muscle fatigue model, $\int_{0}^{t} {\Gamma (u, \theta, \dot{\theta}, \ddot{\theta})} du$ is  the joint momentum. This is the most important difference between dynamic muscle fatigue model and static muscle fatigue model. In static situation the joint torque is a constant, and with time goes by the joint momentum is a linear function. In dynamic muscle fatigue model the joint torque is changing with joint angle and time. The joint momentum is a non-linear function. Figure~\ref{fig:Momentum}(a) is the torque of elbow joint in a cyclic motion of 2 seconds and there are a part of positive torque and a part of negative torque. We consider the positive torque as a result of the effect of angonistic muscle and the negative torque as a result of antagonistic muscle. Figure~\ref{fig:Momentum} presented joint torque and momentum evolution of two groups of muscles during 2 seconds. \\
%
\underline{Experimental results:}
The experiment part is an implementation and verification of above mentioned experiment design. At first stage we just measure one person to test its operability and feasibility. A large number of tests will be carried out in the future stage.

\noindent \textbf{Experiment result for one person:}
A male subject ($H=$188cm, $M=$80Kg) took part in the presented experiment. Push and Pull torque of lower arm are measured for the different operation times of $t$ in $\{0,1,2,3,4,5\}$ minutes. The experiment results are presented in Table~\ref{tab:Experiment_data_table}.\\

\vspace{-8mm}
\begin{table}[h!]
\begin{center}
	\begin{tabular}{|c|c|c|c|c|c|c|}
		\hline
		$\Gamma_{cem}[N \cdot m]$ &	0 min   & 1 min  & 2 min   & 3 min   & 4 min   & 5 min\\
		\hline
		Push                 			&	31.46   & 30.08  & 28.07   & 29.33   & 26.32   & 26.82\\
		\hline
		Pull                 			&	31.71   & 28.33  & 22.94   & 22.31   & 20.05   & 18.67\\	                
		\hline
	\end{tabular}
	\caption{\label{tab:Experiment_data_table} Current exertable maximum joint torque for push and pull action}
\end{center}
\end{table}

\vspace{-8mm}
\noindent \textbf{Fatigue rate parameter $k$ evaluation:}
The parameter $k$ represents fatigue rate and it depends on individual person itself. To evaluate the parameter $k$ of our model, we suppose $k$ is constant. The following Eq.~\eqref{eq:Parameter_k} which is deduced from Eq.~\eqref{eq:fatigue6} is used to calculate $k_i$ with the help of using the experiment measurement of $\Gamma_{cem_i}$ for each operation.
\begin{equation}
	k_i = - \ln \left( \frac{\Gamma_{\cem}(t)}{\Gamma_{\MVC}} \right) / \int_{0}^{t_i} \Gamma (u, \theta, \dot{\theta}, \ddot{\theta})du
  \label{eq:Parameter_k}
\end{equation}
For $t=1,2,3,4,5$ minutes, the agonistic and antagonistic muscle group fatigue rate were evaluated as follows: 
$$k_{agonist}= [0.13, 0.17, 0.07, 0.13, 0.09] [{\rm min}^{-1}]$$
$$k_{antagonist}= [19.56, 28.07, 20.32, 19.86, 18.36] [{\rm min}^{-1}]$$
\begin{floatingfigure}[r]{50mm}
	\centering
		{\includegraphics[width=50mm]{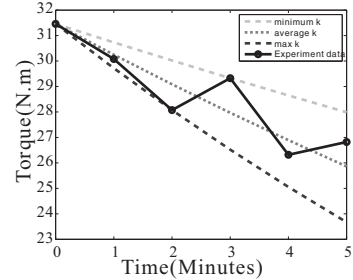}}\hfill
	\caption{\label{fig:k_fitting}Theoretical evolution of $\Gamma_{cem}$ and experiment data using different values of $k_{agonist}$}
\end{floatingfigure}
\vspace{-5mm}
In Ma~\cite{Ma2009Thesis}, the values of $k$ obtained are around 0.87, so $k_{agonist}$ is a realistic value due to that the fact the blood circulation is better during dynamic motions. Conversely, $k_{antagonist}$ seems to be too high. In fact, due to the co-contraction activities influence, the torque of the antagonistic muscle group is higher than the results computed by the dynamic model. To characterize $k_{antagonist}$ more precisely, another experimental measurement is necessary to make the same motion with a pulley based system that inverse the gravity force. Because of the measurement errors of forces, the calculated $k$ is not exactly the same for each time $t$. 
To evaluate the confidence of the fatigue rate parameter $k$, with the minimum, average and maximum values of $k_{agonist}$, $\Gamma_{cem}$ is evaluated seperately and compared with the experimental measurements in Fig.~\ref{fig:k_fitting}. It seems that the first two experimental measurements overestimate $k_{agonist}$. This means that we have to wait three minutes to have a good evaluation of the muscle fatigue properties. In fact, we can consider the force capacity of one muscle group can increase in the beginning of the activity as a warming-up period of the muscle. As only one person participated the experiment, the conclusion cannot be generalized but we have obtained interesting informations. The test will be done for a representivie number of participants in the future works.

%% file: Discussion.tex
\section{Conclusion and Perspectives}
In this paper, a new muscle fatigue model for dynamic motions is presented. Thanks to the robotic method, dynamic factors have been introduced to characterize a new dynamic muscle fatigue model from the joint level. This model can be explained theoretically. Meanwhile, an experiment has been designed to validate it. This model could demonstrate the potential for predicting muscle fatigue in dynamic motions. The limit of this work is that it still lacks experimental validation for more participants. In the future, validations of experiments for a number of participants will be carried out.

%% file: ark2012-template.bbl
\begin{thebibliography}{10}
\providecommand{\url}[1]{{#1}}
\providecommand{\urlprefix}{URL }
\expandafter\ifx\csname urlstyle\endcsname\relax
  \providecommand{\doi}[1]{DOI~\discretionary{}{}{}#1}\else
  \providecommand{\doi}{DOI~\discretionary{}{}{}\begingroup
  \urlstyle{rm}\Url}\fi

\bibitem{AitHaddou2000}
Ait-Haddou, R., Binding, P., Herzog, W.: Theoretical considerations on
  cocontraction of sets of agonistic and antagonistic muscles.
\newblock Journal of Biomechanics \textbf{33}(9), 1105 -- 1111 (2000)

\bibitem{Chaffin1999}
Chaffin, D.B., Andersson, G.B.J., Martin, B.J.: Occupational Biomechanics, 3rd
  Edition.
\newblock Wiley-Interscience (1999)

\bibitem{Giat1993}
Giat, Y., Mizrahi, J., Levy, M.: A musculotendon model of the fatigue profiles
  of paralyzed quadriceps muscle under fes.
\newblock IEEE Transactions On Biomedical Engineering \textbf{40(7)}, 664--674
  (1993)

\bibitem{Hill1983}
Hill A, V.: The heat of shortening and dynamic constants of muscle.
\newblock Biological Sciences \textbf{126}, 136--195 (1983)

\bibitem{Karp2001}
Karp, J.R.: Muscle fiber types and training.
\newblock Strength and Conditioning Journal \textbf{23}(5), 21 (2001)

\bibitem{Khalil2010}
Khalil, W.: Modeling and control of manipulators.
\newblock Ecole Central de Nantes (2009-2010)

\bibitem{Kumar2008}
Kumar, R., Kumar, S.: Musculoskeletal risk factors in cleaning occupation: a
  literature review.
\newblock International Journal of Industrial Ergonomics \textbf{38}, 158--170
  (2008)

\bibitem{Liu2002}
Liu, J., Brown, R., Yue, G.: A dynamical model of muscle activation, fatigue,
  and recovery.
\newblock Biophysical Journal \textbf{82(5)}, 2344--2359 (2002)

\bibitem{Ma2009Thesis}
MA, L.: Contributions pour l'analyse ergonomique de mannequins virtuels.
\newblock Ph.D. thesis, Ecole Central de Nantes (2009)

\bibitem{Westgaard2000}
Westgaard, R.H.: Work-related musculoskeletal complaints: some ergonomics
  challenges upon the start of a new century.
\newblock Applied Ergonomics \textbf{31}, 569--580 (2000)

\bibitem{Wexler1997}
Wexler, A.S., Ding, J., Binder-Macleod, S.A.: A mathematical model that
  predicts skeletal muscle force.
\newblock IEEE Transactions On Biomedical Engineering \textbf{44(5)}, 337--348
  (1997)

\end{thebibliography}
